\documentclass[letterpaper]{article} 
\usepackage{aaai2027}  
\usepackage[hyphens]{url}  
\usepackage{graphicx} 
\usepackage{natbib}  
\usepackage{caption} 
\usepackage{algorithm}
\usepackage{algorithmic}
\usepackage{amsmath}
\usepackage{booktabs}
\usepackage{xcolor}
\usepackage{colortbl}
\definecolor{ourshade}{gray}{0.93}

\usepackage{tabularx}

\title{CoEvo: Oracle-Grounded Self-Evolution of a Single Model\\for Multi-Step Causal Reasoning}
\author{Jian Zhang, Bingyi Wang, Yizhi Liu}
\affiliations{Zhejiang University}

\nocopyright
\begin{document}

\maketitle

\begin{abstract}
Multi-step causal reasoning requires chaining inferences where each step constrains the next. An early error propagates silently, and a correct answer reached via flawed logic evades outcome-level detection. In specialized domains, teacher LLMs err on intermediate steps, safety constraints restrict cloud distillation, and shifting conditions demand adaptation, leaving self-evolution as the practical route. Naive self-evolution can collapse: outcome-only rewards let the model exploit distributional shortcuts, and weak self-evaluation reinforces spurious paths into stable failure patterns. We exploit a key asymmetry: generating a correct chain is hard, but verifying a single step is easy. Many high-stakes domains admit a deterministic, queryable oracle, a physics simulator or rule engine over codified constraints. It checks asserted steps without teacher-level ability and abstains beyond its rules; it can check what the model asserts, never replace it. This enables CoEvo, an oracle-grounded self-evolution framework where a single model alternates between Proposer and Solver. As Solver, the model generates competing chains; intra-group debate exposes disagreement steps, a proxy for the capability boundary, and the oracle adjudicates them into process-level supervision. As Proposer, the same model constructs progressively harder scenarios inside oracle constraints, steering the curriculum toward deep multi-hop chains. Both roles are updated jointly, so training pressure co-evolves with the model. On industrial, clinical, and legal multi-step causal reasoning benchmarks, CoEvo enables an 8B LLM to sustain self-evolution, surpassing distillation baselines and the strongest proprietary reference on path correctness ($82.1\%$ vs.\ $71.4\%$). The trained model generalizes to unseen categories and systems, preserving root-cause accuracy.
\end{abstract}

\section{Introduction}

In industrial fault diagnosis, clinical diagnosis, and legal analysis \citep{appliedenergy2024fault,singhal2023large,guha2023legalbench}, large language models are increasingly asked to perform multi-step causal reasoning: tracing an observed effect back to its cause through a chain of inferences, each step depending on the one before. These settings run on proprietary operational records, patient data, and case files, so open, locally deployable models are the practical choice for privacy, latency, and controllability. Yet deployable scale is precisely where this capability is missing: gains from chained intermediate reasoning emerge reliably only in far larger models \citep{wei2022chain}. Our goal is to train a locally deployable LLM that carries out this reasoning reliably.

The obvious route is knowledge distillation \citep{hinton2015distilling}: the deployable model, now a student of a stronger teacher, inherits reasoning it could not reach alone. In specialized domains this route fails for three reasons. The teacher's process itself is not fully correct, and not by accident: tracing effects back to causes forces a choice among competing explanations at almost every step, and even the strongest LLMs err systematically on exactly such steps \citep{jin2023cladder}; imitation then transfers the flawed steps along with the fluent form \citep{gudibande2024imitation,turpin2023language}. Confidentiality forbids routing proprietary data through an external teacher. And deployed conditions keep drifting after any fixed supervision ends. Supervising the student's own rollouts instead, as on-policy distillation does \citep{gu2024minillm,agarwal2024onpolicy}, corrects the student where it actually fails, but at the price of a deeper dependence: the teacher must now judge reliably at every state the student visits, throughout training. What remains is for the deployable model to supervise its own trajectories: self-evolution \citep{zelikman2022star,yuan2024selfrewarding,zhao2025absolute,huang2025rzero}.

Self-evolution asks the model to improve by finding and repairing its own deficiencies. Two obstacles stand in the way. The first obstacle is that the judge is the party being judged. Naive self-evolution collapses here: the model must certify its own reasoning while its self-evaluation lags behind its generation, and self-correction rarely persists without reliable external feedback \citep{huang2024large,kamoi2024when}. On step-dependent chains the cost compounds: each step constrains the next \citep{wei2022chain}, an early error propagates silently, and a chain can reach a correct conclusion through flawed logic \citep{uesato2022solving,lightman2024lets,turpin2023language}. Outcome-only rewards grant such chains full credit, so the model learns distributional shortcuts rather than causal structure \citep{guo2025deepseekr1,geirhos2020shortcut}, and errors it cannot see are re-sampled and re-rewarded into stable failure patterns. Nor does adding judges of the same origin help: voters in a two-model self-play loop are fine-tuned from one base model and err together \citep{huang2025rzero}, and verifiers learned from model-generated traces can inherit the very blind spots they are meant to detect \citep{zhang2025lessons}. Across these designs, step-level judgment stays in model space. The second obstacle is finding what to repair. Improvement concentrates on tasks near the current capability frontier \citep{bengio2009curriculum}, a frontier that is model-specific and moves with every update, so no pre-existing task collection tracks it: training tasks must be generated continually, aimed at where the model currently fails.

But step-dependent reasoning has one feature that works in self-evolution's favor: chains grow by composition. A model that solves depth-$k$ chains is close to solving depth-$k{+}1$, typically one junction short: one that already traces a stuck valve to a cold coil needs only one further junction, from the cold coil to the low supply-air temperature it causes. Difficulty therefore comes with a built-in staircase, and the staircase is model-independent: how hard a task is can be read off its structure, the depth of the chain and the kinds of constraints in play, without consulting the model. What the staircase does not do is build itself: the next step must be laid as the model climbs, and at the model's pace. Taken together, whether self-evolution works comes down to three conditions: every asserted step needs a verdict it can trust (the supervision requirement); the model's current failures must be exposed (the localization requirement); and the training tasks must adapt as the model improves, well-posed, difficulty-graded, and co-evolving with it (the adaptation requirement).

All three requirements can be met around a single external anchor. Our key observation is that generating a correct chain is hard, but checking a single asserted step is easy, and checking needs no teacher: many high-stakes domains already maintain a queryable form of their own knowledge, a codified knowledge base, a physics simulator, a rule engine, an executable sandbox, against which an asserted step can be checked mechanically \citep{pan2023logiclm,zhao2025absolute}: fault-propagation constraints in the industrial domain, guideline-derived criteria in the clinical, statutory elements in the legal. We call this arbiter an oracle. Its verdicts are repeatable and independent of the model being trained, which meets the supervision requirement. It cannot search the hypothesis space or read free-form observations, so it cannot find where the model fails; but disagreement among the model's own competing solutions can, which is how the localization requirement will be met. And because the oracle can certify a task and grade its structural difficulty before training, task generation can be anchored on it, which is how the adaptation requirement will be met. Yet the oracle can only check what the model asserts, never replace it. Self-evolution built this way is oracle-grounded rather than unsupervised: no teacher LLM and no demonstrations, but a real dependence on codified domain knowledge.

We therefore propose \textbf{CoEvo}, a self-evolution framework that puts this oracle to work in one training loop meeting all three requirements. Role-conditioned prompts switch a single policy between Proposer and Solver; alternation is in data generation only, and both role batches are optimized jointly in each training iteration. The Proposer instantiates each scenario as an executable instance in the oracle's schema; the oracle certifies its constraint consistency and fixes the hidden ground-truth answer, the anchor, before training. The Solver samples competing chains, which the oracle screens step by step for local validity. Steps where chains commit to incompatible assertions go to an intra-group debate: debate decides nothing itself, but forces each side to name the constraint and instance query that would separate the branches, which the oracle then executes against the certified instance. Verdicts feed a process-level reward; disputes the oracle cannot separate contribute no reward and, across epochs, return as scenarios that expose the missing evidence, while disagreement statistics redirect proposing. The curriculum climbs from single-hop toward deep multi-hop tiers on a structural axis of chain depth and constraint families; each deeper tier extends mastered transitions by one junction. The axis is model-independent, while advancement follows the model's measured accuracy and disagreement. No judge role is trained anywhere in this loop: disagreement locates, the oracle settles, and proposing adapts to what adjudication reveals. This paper makes the following contributions:
\begin{itemize}
\item We introduce reliable supervision into self-evolution without importing the model's own bias: reward is issued only on verdicts of the deterministic domain oracle, never on the model's own judgment.
\item We enable the model to expose and repair its own weaknesses: contradictions among its attempts locate its weakest steps, and new tasks keep targeting them, so ability and curriculum evolve together.
\item We validate CoEvo on industrial, clinical, and legal causal reasoning benchmarks: an 8B model sustains self-evolution over eight iterations, surpasses proprietary and distillation baselines on path correctness and on unseen-category generalization in the industrial and clinical domains, and matches the strongest reference on macro-average root-cause accuracy.
\end{itemize}

\section{Related Work}

\subsection{From Distillation to Self-Evolution}

Knowledge distillation transfers capability from a stronger teacher \citep{hinton2015distilling}, and on-policy distillation supervises the student's own rollouts \citep{gu2024minillm,agarwal2024onpolicy}; both presuppose the reliable teacher specialized domains lack \citep{gudibande2024imitation}. Self-improvement therefore internalizes supervision, from answer-checked rationales and self-scored outputs to verifiers learned for intermediate steps \citep{zelikman2022star,yuan2024selfrewarding,wang2024mathshepherd}. Reinforcement learning with verifiable rewards adds an external check, though typically at the final answer \citep{guo2025deepseekr1}. Co-evolving frameworks extend self-evolution to task generation. R-Zero trains a separate Challenger to pose questions near the Solver's ability edge, with the Solver pseudo-labeled by answer-consistency voting and difficulty read off the Solver's own uncertainty \citep{huang2025rzero}; nothing in the loop judges an individual reasoning step. Absolute Zero comes closest to our setting: a single model proposes and solves code tasks that an executor validates and verifies \citep{zhao2025absolute}; its checks address a program's behavior and final answer, and its curriculum follows estimated learning progress. Multi-Agent Evolve adds a Judge role drawn from the same LLM \citep{chen2025mae}. These frameworks advance the task-generation side of the loop, but the two obstacles of Section~1 remain: whether an asserted step is right, and where the model currently fails, still rest on the model's own judgment.

\subsection{Task Generation and Curricula}

Automatic task generation ranges from heuristic filtering of a model's own instructions \citep{wang2023selfinstruct} to asymmetric self-play, where a goal-setting agent supplies the learner's curriculum \citep{sukhbaatar2018intrinsic,bengio2009curriculum}. In self-evolving LLMs the two halves of task quality diverge. Validity can be checked externally, as executors do for self-contained programs \citep{zhao2025absolute}; difficulty grading, however, stays model-relative across these systems, from vote splits and estimated learning progress \citep{huang2025rzero,zhao2025absolute} to curriculum policies learned online from the learner's own advantage signal \citep{chen2025sec}. A curriculum for step-dependent reasoning must therefore satisfy the adaptation requirement: tasks certified and graded outside the generator, supplied at the learner's pace.

\subsection{Debate and Multi-Agent Reasoning}

Debate entered the field as a scalable-oversight proposal in which a judge, originally a human, decides between adversarial agents \citep{irving2018debate}. Its LLM instantiations improve answers at inference time, converging by consensus or persuading a judge model \citep{du2024improving,khan2024debating}, and debate has also been trained through self-play, making arguments easier for a judge model to evaluate \citep{arnesen2024debate}. In these systems the exchange is discarded once the answer is produced, and arbitration rests with a vote or a judge model. Debate surfaces the steps where competing attempts collide, exactly what the localization requirement needs at training time, with adjudication left to an arbiter outside the debaters.

\paragraph{Summary.}
Each of these lines advances one part of the loop: co-evolving frameworks generate tasks and check final answers, curricula pace difficulty, debate exposes disagreement. None of them, however, suffices to support reliable self-evolution over step-dependent chains, which requires step verdicts the model cannot bias, weaknesses located as they emerge, and a task supply that adapts under external quality control. \textbf{CoEvo} is built around exactly these three requirements: a domain oracle holds every correctness judgment outside the model, while disagreement among the model's own attempts locates weaknesses and steers new tasks. Correctness stays outside the model; localization stays model-relative by design.

\section{Method}
\begin{figure*}[t]
\centering
\includegraphics[width=0.95\textwidth]{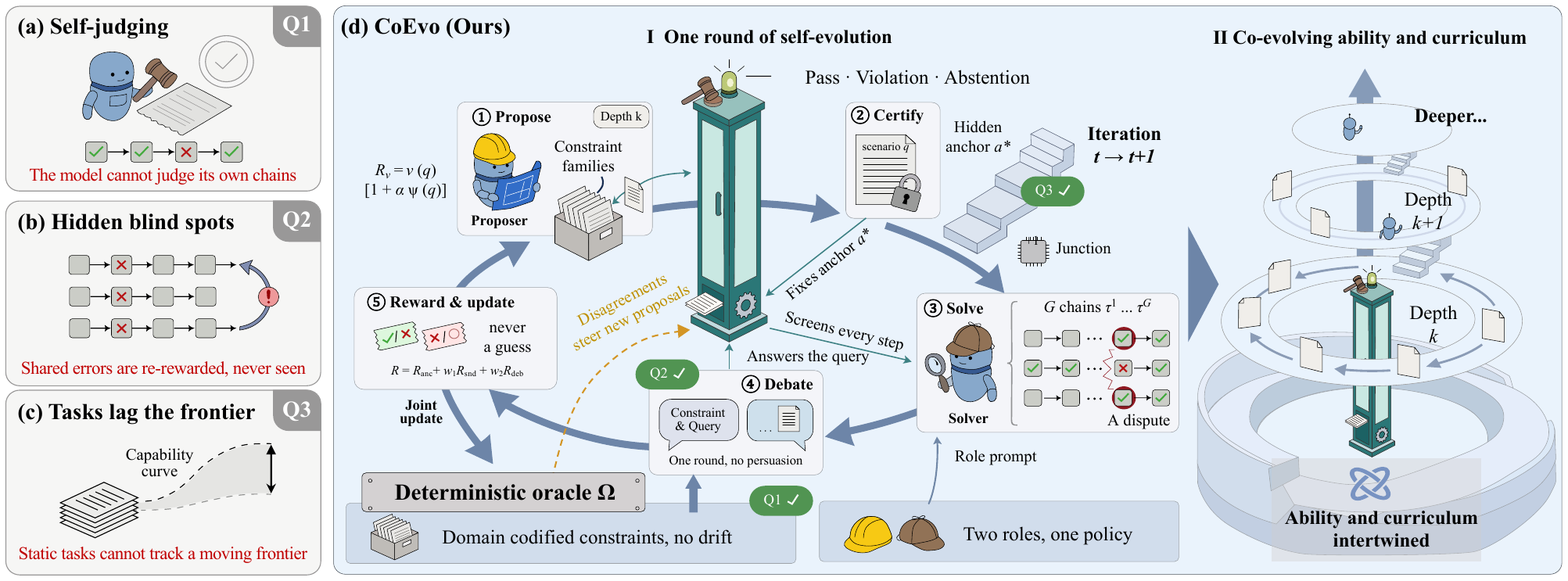}
\caption{(a--c) Failure modes of unverified self-evolution; (d) the CoEvo training loop.}
\label{fig:overview}
\end{figure*}

\subsection{Overview and Problem Formulation}

CoEvo trains a single locally deployable policy $\pi_\theta$ to perform multi-step causal reasoning through self-evolution: the model constructs its own training tasks as Proposer, solves them as Solver, and receives process-level supervision from a deterministic oracle that adjudicates individual reasoning steps (Figure~\ref{fig:overview}); no second model and no teacher are involved.

The division of labor rests on a directional asymmetry underlying Section~1's observation. Solving is backward: from observed symptoms the model must discriminate candidate causes whose symptom signatures overlap. Proposing is forward: from one fixed cause it expands consequences along codified constraint families, and the branching factor collapses. Hence the architecture: the model is disqualified as a judge of its own answers, yet qualified as a constructor under an anchor the oracle certifies, so Proposer and Solver can be two prompts over one policy while every verdict stays outside it; role separation refers to prompts, rollouts, and objectives, not to parameters. The assumed forward advantage is measured in Section~4.1.

CoEvo applies external checks at three points of the loop: step screening (Section~3.2), task certification (Section~3.3), and dispute adjudication (Section~3.4); disagreement statistics then steer the next round of proposing. One principle runs through the loop, the warrant principle: only an oracle certificate or verdict can contribute reward, in either direction. Model disagreement decides where supervision is allocated, never whether an assertion is correct.

A task is a scenario $q$ describing observations of a system governed by domain constraints, paired with a ground-truth anchor $a^{\ast}$ fixed at proposal time. Solving $q$ means producing a reasoning chain $\tau=(u_1,\ldots,u_K)$: each step $u_k$ asserts one causal transition over the scenario schema, the vocabulary of entities, states, and constraints the oracle exposes, together with its premises, and the chain terminates in a conclusion checked against $a^{\ast}$. Steps are emitted in a structured format, so every asserted transition is a discrete, queryable object; step dependence arises because the premises of $u_k$ must be established by the observations or by earlier steps. We use the term \emph{causal transition} throughout; legal dependence is statutory conditioning (Section~4.1). Task difficulty grows along two measurable axes: chain depth and the number of active constraint families.

\subsection{Deterministic Oracle Interface}

The oracle $\Omega$ is a pluggable, deterministic, queryable adjudicator in any of the forms of Section~1; our three domains all use a rule engine (Section~4.1). Queried with an asserted transition and its stated premises, it returns a step verdict
\begin{equation}
\Omega(u \mid q) \in \{\textit{pass}, \textit{violation}, \textit{abstention}\}.
\label{eq:verdict}
\end{equation}
Here $q$ is the exposed scenario the Solver sees, not the hidden certified instance. A transition receives \textit{pass} when it is consistent with the applicable domain constraints under its premises, \textit{violation} when it contradicts one, and \textit{abstention} when it falls outside the oracle's competence; an abstention carries a non-reward reason code, under-specified query or out of coverage, used in Sections~3.3--3.4. Verdicts come from constraint evaluation, not learned inference: no training, no drift, identical queries yield identical verdicts; the oracle is only as correct and complete as its codified constraints. The absence of a violation does not certify a chain as correct. Three notions of correctness are kept apart: \emph{local validity}, whether a transition holds under its stated premises, checked by screening, the application of Eq.~\eqref{eq:verdict} to every asserted step; \emph{premise consistency}, whether those premises hold in the scenario, decidable because the oracle retains the hidden instance (Section~3.3); and \emph{global correctness}, whether the chain reaches the true root cause, scored by the anchor check.

The oracle is not a teacher: it generates no trajectories and cannot prescribe what the model should assert. An abstention earns no reward rather than a false one; an under-specified query can be repaired through debate (Section~3.4), while out-of-coverage content is never converted into supervision. A pass certifies one assertion under its premises, leaving comparisons among locally valid chains to debate. Beyond judging, the oracle exposes its constraint families to the Proposer (Section~3.3); its three formal calls are specified in the technical appendix. CoEvo thus trades teacher demonstrations for an executable domain specification: it applies where constraints are codified but expert trajectories are scarce or restricted.

\subsection{Proposer Side: Oracle-Anchored Curriculum}

As Proposer, the model does not write scenarios free-form: it instantiates an executable instance in the oracle's schema, a latent causal chain composed from the exposed constraint families, and selects a candidate anchor $a^{\ast}$; the oracle certifies the instance only if the derived observables identify that anchor, and fixes it as the hidden ground truth. Derived observables are rendered into the natural-language scenario through deterministic templates; a rule-based parser checks the round trip, requiring exactly the derived observables and no trace of the anchor, so the faithfulness check is deterministic, not a second model judge. Proposals failing certification or faithfulness earn zero reward; validity is a property of the certified instance. The oracle could enumerate valid instances but cannot judge which are worth training on; choosing what to pose is the Proposer's responsibility. The Proposer is rewarded by
\begin{equation}
R_{P}(q) \;=\; v(q)\,\bigl[\,1 + \alpha\,\psi(q)\,\bigr],
\label{eq:proposerreward}
\end{equation}
where $v(q)\in\{0,1\}$ is the oracle's validity gate and $\psi(q)$ is the fraction of the certified instance's transitions that are disputed and later separated by the oracle (Section~3.4), so it rewards scenarios that generate many adjudicable disputes rather than one easy one; it is zero when nothing is disputed. Disagreement is measured from the Solver's behavior and adjudicated by the oracle, so the reward targets the capability boundary without asking the model to judge difficulty. Since certification already excludes under-determined scenarios and $\psi(q)$ counts only disputes the oracle later separates, unseparated disputes add no bonus: a proposal earns beyond its validity gate only through verdicts the oracle actually issued. Disagreement also keeps the group-relative learning signal alive: a scenario on which all $G$ chains agree teaches little.

The difficulty axis is structural rather than model-judged: each scenario belongs to a tier defined by chain depth and active constraint families. The curriculum advances toward deeper tiers when anchor accuracy at the current tier exceeds a threshold and disagreement there falls, with violation rate, audit error, and output diversity held within fixed bounds (appendix); since the Proposer is rewarded for residual disagreement, a tier advances exactly when disagreement-seeking proposals no longer split the group, and dispute statistics redirect proposing toward high-dispute constraint families. When the tier advances, the Proposer composes each deeper scenario by extending, by one new junction, transitions the Solver has repeatedly defended under screening and adjudication, typically from a high-dispute family: a depth-$k{+}1$ chain reuses as premises the components mastered at depth $k$ \citep{bengio2009curriculum}, the component set defined online by adjudication history rather than a fixed skill library. Section~4.4 isolates this extension rule against same-tier proposing without it.

\subsection{Solver Side: Group Rollouts and Debate}

For each scenario the Solver samples $G$ competing chains $\tau^{1},\ldots,\tau^{G}$, and the oracle screens every step under Eq.~\eqref{eq:verdict}, computing the soundness term $R_{\text{snd}}$ of Eq.~\eqref{eq:solverreward}: a violating step is penalized whether or not any chain disagrees, so per-step checking and dispute-only adjudication are complementary layers, not alternatives. Screening also enforces provenance: every premise must cite, by identifier, an exposed observation or a predecessor that itself passed screening, so unsupported premises cannot propagate.

Disagreement is defined over asserted claims, not surface positions: all chains instantiate transitions from one scenario schema, so no sequence alignment is needed and paraphrases do not register. A dispute is a pair of committed assertions that the schema marks incompatible,
\begin{equation}
\mathcal{D}(q) \;=\; \bigl\{ (u,u') : u \in \tau^{i},\; u' \in \tau^{j},\; \Gamma_{q}(u,u')=1 \bigr\},
\label{eq:disputed}
\end{equation}
where the incompatibility relation $\Gamma_{q}$ holds in exactly two schema-defined cases: the assertions assign mutually exclusive states to one entity, or, under the certified single-anchor structure (a property of certified tasks, not open diagnosis), they attribute one observable to competing causes. Locating a dispute is thus mechanical, not model-judged. The disputed transitions are a proxy for the capability boundary, junctions on which the model's own attempts genuinely diverge; a proxy, not a definition: it misses errors shared by all sampled chains. Screening and the anchor check catch some of these, and a low-rate random audit of consensus steps estimates the residual error; the audit is reported in Section~4.4 and never enters the reward.

Debate is held only on disputed transitions, in a narrow sense: one round, oracle-settled, no persuasion. For each pair in $\mathcal{D}(q)$, the chain asserting one side challenges the other and must state the constraint and premises its assertion relies on; the affirming chain defends its own. Since each step already states its premises, the debate's job is to select between branches: the challenger names the constraint and the instance query that would separate them, and the oracle runs it against the certified instance, a check screening never performs. The same route repairs an under-specified-query abstention (Section~3.2), whose verdict then counts in the debate term. Instance-grounded checks cost more than rule-local screening and are budgeted, routed to disputed steps where rule-local screening is silent; Section~4.4 varies this routing under a fixed budget, and a worked industrial example appears in the technical appendix.

All $G$ chains are sampled from one policy and share its blind spots, so debate carries no adjudication authority: it locates disputes, only the oracle settles them, and the exchange cannot be won by persuasion, majority, or collusion. A dispute beyond the codified constraints, or with no admissible query available, contributes no reward rather than a guessed verdict. Although certification retains the full instance, the instance-query call accepts only predicates over admissible evidence: a query naming an unexposed variable is rejected, the dispute scores zero, no chain is penalized for evidence it could not see, and the transition is queued for re-instantiation in a new certified scenario that exposes the discriminating observation without revealing the anchor; supervision resumes only there. Un-re-posed disputes enter the disagreement statistics (unseparated rate in the appendix); Section~4.4 isolates re-instantiation against statistics-only steering. Debate transcripts are discarded after the update. Disagreement locates, the oracle settles.

\begin{table*}[t]
\centering
\small
\setlength{\tabcolsep}{5pt}
\begin{tabularx}{\textwidth}{Xcccccccc}
\toprule
& \multicolumn{2}{c}{\textbf{Industrial}} & \multicolumn{2}{c}{\textbf{Clinical}} & \multicolumn{2}{c}{\textbf{Legal}} & \multicolumn{2}{c}{\textbf{Average}} \\
\cmidrule(lr){2-3} \cmidrule(lr){4-5} \cmidrule(lr){6-7} \cmidrule(lr){8-9}
Method & Path & Acc & Path & Acc & Path & Acc & Path & Acc \\
\midrule
\multicolumn{9}{c}{\textbf{References (zero-shot)}} \\
\midrule
Qwen3-8B (base)                 & 23.01 & 32.32 & 20.52 & 35.87 & 35.25 & 44.60 & 26.26 & 37.60 \\
Claude-Sonnet-4.6               & 74.68 & 87.54 & 59.91 & 84.09 & 79.50 & 90.29 & 71.36 & 87.31 \\
Qwen3.7-Max                     & 72.96 & 79.27 & 56.91 & 80.41 & 73.38 & 79.50 & 67.75 & 79.73 \\
\midrule
\multicolumn{9}{c}{\textbf{Distillation (Qwen3-8B)}} \\
\midrule
SFT on certified data           & 66.06 & 78.51 & 52.25 & 69.87 & 49.64 & 61.87 & 55.98 & 70.08 \\
Teacher-trajectory distillation \citep{agarwal2024onpolicy} & 64.21 & 79.20 & 50.66 & 70.99 & 47.84 & 63.31 & 54.24 & 71.17 \\
\midrule
\multicolumn{9}{c}{\textbf{Self-evolution (Qwen3-8B)}} \\
\midrule
Outcome-only self-evolution     & 37.87 & 79.68 & 28.64 & 74.47 & 30.22 & 65.11 & 32.24 & 73.09 \\
\rowcolor{ourshade}
\textbf{CoEvo (full)}           & \textbf{92.67} & \textbf{95.00} & \textbf{72.93} & \textbf{81.14} & \textbf{80.58} & \textbf{87.05} & \textbf{82.06} & \textbf{87.73} \\
\bottomrule
\end{tabularx}
\caption{Main results (\%): reasoning path correctness (Path) and root-cause/conclusion accuracy (Acc); Average is the unweighted mean across domains. Best 8B-scale result in bold.}
\label{tab:main}
\end{table*}

Each chain is scored by
\begin{equation}
R(\tau^{i}) \;=\; R_{\text{anc}}(\tau^{i}) \;+\; w_{1}\,R_{\text{snd}}(\tau^{i}) \;+\; w_{2}\,R_{\text{deb}}(\tau^{i}),
\label{eq:solverreward}
\end{equation}
where $R_{\text{anc}}$ is positive only when the conclusion matches the anchor $a^{\ast}$ and the chain covers a sufficient support set for the certified instance, one the oracle accepts as establishing the anchor rather than one fixed template chain, so a bare conclusion with a few safe steps cannot collect the anchor term; $R_{\text{snd}}$ averages screening verdicts over the chain's steps, scoring \textit{pass} $+1$, \textit{violation} $-1$, and \textit{abstention} $0$ so that content outside oracle competence earns no credit; and $R_{\text{deb}}$ scores adjudicated outcomes on disputed transitions, $+1$ when a defended assertion is upheld, $-1$ when overturned, $0$ for unseparated disputes, rewarding assertion correctness only, not query formulation. Each term draws only on verdicts the oracle issued, the warrant principle applied to the Solver; a disputed step is scored on two axes, $R_{\text{snd}}$ for local validity and $R_{\text{deb}}$ for instance-level adjudication. The anchor term alone would reproduce the outcome-only failure mode of Section~1; the soundness and debate terms are what make the reward process-level.

\subsection{Joint Optimization}

Both roles are updated in one joint update:
\begin{equation}
\mathcal{L}(\theta) \;=\; (1-\lambda)\,\mathcal{L}_{S}(\theta) \;+\; \lambda\,\mathcal{L}_{P}(\theta) \;+\; \beta\,\mathrm{KL}\bigl(\pi_\theta \,\|\, \pi_{\text{ref}}\bigr),
\label{eq:jointloss}
\end{equation}
where $\mathcal{L}_{S}$ and $\mathcal{L}_{P}$ are policy-gradient losses under Eqs.~\eqref{eq:solverreward} and~\eqref{eq:proposerreward}, with advantages normalized over each scenario's $G$ chains and each proposal batch, respectively; $\pi_{\text{ref}}$ is the initial policy. Any group-relative optimizer suffices \citep{shao2024deepseekmath}. Joint parameter updates couple the roles; the co-evolutionary feedback itself is the adjudication-conditioned shift in the next round's proposal distribution, so the Solver's current disagreement shapes subsequent proposals without a separate curriculum model. Parameter sharing has a structural basis: every transition the Proposer expands forward is one the Solver must recognize backward; whether forward practice strengthens backward solving is tested against separately trained roles in Section~4.4.

The full epoch pseudocode, role prompts, tier schedule, and reward coefficients are in the technical appendix. At inference only the trained model is deployed: it answers from the scenario alone, without debate or oracle access.

\section{Experiments}

\subsection{Experimental Setup}
\label{sec:setup}

\paragraph{Benchmarks and protocol.}
We evaluate on three domains: industrial fault diagnosis on AHU cases, the LBNL dual-duct corpus \citep{granderson2022lbnl} split 80/20 with the full LBNL single-duct and ASHRAE RP-1312 sets \citep{wen2011rp1312} added to the test side as cross-system suites; clinical diagnosis on DDXPlus under its official split \citep{tchango2022ddxplus}, whose codified generative knowledge base supports a matched rule-engine oracle (guideline-codified diagnosis, not open clinical practice); and legal reasoning on MSLR \citep{yu2025mslr}, split 80/20 by case. The three domains instantiate one operational task, multi-step attribution over checkable transitions, with physical, guideline-codified, and normative dependence respectively. Unseen-category evaluation holds out industrial fault categories and system types and clinical pathologies. Case splits precede training, proposed instances are deduplicated against evaluation cases, and held-out categories' rule instantiations are withheld from the training-time oracle, so unseen categories are new compositions within the shared schema, not new rule families.

The oracle is instantiated per domain as a rule engine compiled from the same knowledge that governs the benchmark cases, so proposed and benchmark cases share one schema; per-domain audits (coverage, abstention, certification false positives, faithfulness, expert spot-checks) are in the appendix. What it supplies is executable correctness against the benchmark's knowledge base, not real-world clinical or legal validity. The deployable model is Qwen3-8B; a 14B variant appears in the appendix. Building the three oracles took $90$/$186$/$58$ rules and $40$/$62$/$28$ expert-hours, and training costs $296$ GPU$\cdot$h on $4\times$NVIDIA H200 (appendix).

\paragraph{Baselines.}
We compare against four groups. \emph{References}: the base Qwen3-8B zero-shot (B1) and two proprietary references, Claude-Sonnet-4.6 and Qwen3.7-Max, zero-shot (B2; GPT-4o in the appendix). \emph{Distillation}: SFT on oracle-validated generated data without self-evolution (B3) and standard teacher-trajectory distillation (B4). \emph{Self-evolution}: naive self-evolution with outcome-only reward, one role, and no oracle (B5). \emph{CoEvo variants} for the mechanism ablations of Section~4.4. All methods share one prompt set, output schema, and evaluation protocol; trained rows share the Qwen3-8B backbone.

\paragraph{Metrics.}
We report four process-level metrics: reasoning path correctness (primary), root-cause accuracy (conclusion accuracy in the clinical and legal domains), per-hop accuracy, and the oracle-judged constraint violation rate. A chain counts as path-correct when its conclusion matches the anchor, every checkable step passes screening with grounded premises, and its checkable steps cover a sufficient support set that independently establishes the anchor, so neither a short chain nor one padded with abstentions can score by omission (appendix). The scorer is frozen before training and validated against method-blinded expert audits ($89.7$--$93\%$ agreement), so Path is not graded by the training loop. Training dynamics are tracked by capability, difficulty-gap, disagreement, and output-diversity curves. All reported scores average five independent evaluation runs. Comparisons are paired at case level with bootstrap confidence intervals and McNemar--Holm tests; self-evolution training uses three seeds, and curves carry variance bands. The base model's directional asymmetry (Section~3.1) is confirmed before training: forward beats backward by $29.5$/$26.4$/$21.2$ points across the three domains; remaining controls are in the technical appendix.

\subsection{Main Results}
\label{sec:main-results}

Table~\ref{tab:main} yields two findings. First, CoEvo exceeds the strongest proprietary reference on path correctness in every domain, by $18.0$ points on industrial, $13.0$ on clinical, and $1.1$ on legal; all path gaps are significant under McNemar--Holm ($p<10^{-18}$) except the legal margin ($p=0.26$). That margin concentrates in deep chains: on the multi-hop subset Claude degrades sharply to $61.9$ while CoEvo holds $72.2$, a gap significant on its own (exact McNemar $p=0.013$); single-hop cases are statistically indistinguishable (appendix). On accuracy CoEvo leads on industrial, Claude on clinical and legal; macro-averages are matched ($87.73$ vs.\ $87.31$). Few-shot and self-consistency prompting lift Claude by at most $3.4$ path points, still below CoEvo (failure modes in the appendix); the strongest distillation baseline trails by $26.1$, and a 14B backbone adds under one point (appendix). Second, the outcome-only baseline confirms the diagnosis of Section~1: its average accuracy ($73.09$) climbs past the SFT level ($70.08$) while its path correctness falls to $32.24$, barely above the untrained base ($26.26$): right conclusions increasingly reached through wrong chains.

\subsection{Self-Evolution Dynamics}
\label{sec:dynamics}

\begin{figure}[!htb]
\centering
\includegraphics[width=\columnwidth]{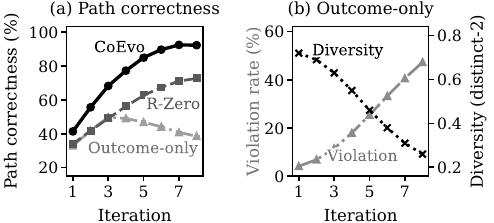}
\caption{Self-evolution dynamics on the industrial probe set (three seeds; $\pm$1 std error bars). (a) Path correctness; the plateau at iterations 7--8 triggers the stopping rule (appendix). (b) Outcome-only baseline: violation rate and output diversity.}
\label{fig:dynamics}
\end{figure}

Figure~\ref{fig:dynamics} tracks eight iterations (three seeds) on a fixed industrial probe set: $600$ certified scenarios, stratified over systems, fault families, and tiers, drawn once from the training-side pool before the first round and excluded from all policy updates; the $3{,}942$-case test set is touched only at final evaluation. CoEvo climbs from $41.3$ to $92.1$ path correctness and holds the plateau, within $0.6$ points of the final test score, indicating no probe overfitting. The outcome-only baseline peaks at iteration~3 and then degrades, violations rising from $4.1\%$ to $47.3\%$ while diversity contracts from $0.72$ to $0.26$: the reinforced-error signature predicted in Section~1, not mere stagnation. Proposed and solved difficulty climb together within $0.4$ tiers, and the disagreement rate falls from $27.1\%$ to $11.8\%$: each tier's disputes are exhausted before the curriculum advances, the intended signature of Eq.~\eqref{eq:proposerreward} (curves in the appendix).

\subsection{Mechanism Ablations}
\label{sec:ablations}

Disputed steps carry $42.5$--$46.1\%$ oracle-judged errors against $7.2$--$10.6\%$ on matched consensus steps, a $4.3$--$5.9\times$ enrichment that validates the disagreement proxy, the consensus rate bounding what it misses (per-domain counts in the appendix); Table~\ref{tab:ablation-core} then isolates one mechanism per row on the industrial domain. Is an external oracle necessary? Mutual voting, returning adjudication to model consensus, costs $13.9$ path points; a learned process-reward model still costs $4.4$: verdicts that drift with the policy supervise worse than executable rules. Do the process terms carry the reward? Anchor-only training drops $8.8$; removing screening costs $3.6$, the debate term $6.3$. Is debate more than a query trigger? Rule-built queries on the same disputed steps cost $4.8$. Is routing efficient? Random queries at equal budget cost $7.7$; dense checking recovers CoEvo within $0.2$ at $3.4\times$ cost. Is the Proposer necessary? Freezing it costs $9.3$, and real-task training with the same oracle rewards but no Proposer costs $7.8$ ($7.5$--$8.6$ across domains): the curriculum, not the process reward alone, carries the gain. Separate role models ($-3.6$) and remaining variants are in the appendix.

\begin{table}[!htb]
\centering
\small
\setlength{\tabcolsep}{3pt}
\begin{tabularx}{\columnwidth}{Xcc}
\toprule
Variant & Path & Acc \\
\midrule
Mutual voting replaces oracle    & 78.79 & 86.96 \\
Learned PRM replaces oracle      & 88.31 & 93.00 \\
Anchor-only reward ($w_1{=}w_2{=}0$) & 83.89 & 91.20 \\
Dense instance checking ($3.4\times$ cost) & 92.44 & 94.93 \\
Frozen Proposer                  & 83.38 & 90.61 \\
Real-task RL, no Proposer        & 84.91 & 90.21 \\
\rowcolor{ourshade}
\textbf{CoEvo (full)}            & \textbf{92.67} & \textbf{95.00} \\
\bottomrule
\end{tabularx}
\caption{Mechanism ablations (\%, industrial, 3,942 cases); PRM: process-reward model; full table in appendix.}
\label{tab:ablation-core}
\end{table}

\subsection{Generalization and Oracle Fidelity}
\label{sec:generalization}

We separate generalization within a system family from the harder RP-1312 cross-system transfer (full table in the technical appendix). On held-out industrial fault categories and clinical pathologies, CoEvo holds $84.36$ and $80.84$ path correctness, $14.2$ and $24.2$ points above the strongest proprietary reference. On RP-1312, where control strategy, sensor distribution, and fault assignment all change and the governing rule instantiations were never used in training, CoEvo holds $85.55$, $16.4$ points above, consistent with transferring causal structure rather than the training oracle's vocabulary. Four checks bound the shared-knowledge-source circularity: blinded expert audits agree with the scorer on $89.7$--$93\%$ of $660$ verdicts and $90.8\%$ on an independent $240$-case path audit; path correctness degrades smoothly, $91.8$/$89.9$/$85.6$ at $90$/$75$/$50\%$ training-oracle coverage with evaluation kept full; the RP-1312 transfer is scored under rules never used in training; and re-scoring the same clinical test set under an independent oracle compiled from the UMLS medical knowledge base \citep{bodenreider2004umls}, built without reference to the training oracle's rules, moves path correctness by $1.1$ points with $91.4\%$ verdict agreement (appendix).

\section{Conclusion}

CoEvo turns a simple asymmetry, that generating a correct causal chain is hard while verifying a single step is easy, into a self-evolution engine: one deployable model alternates between Proposer and Solver, disagreement among competing chains locates the steps to check, a deterministic oracle adjudicates them, and the curriculum extends mastered transitions one junction at a time; no judge role is trained anywhere; at inference the model reasons alone. Across industrial, clinical, and legal benchmarks, an 8B model sustains oracle-grounded self-evolution over eight iterations, surpasses proprietary and distillation baselines on path correctness and on unseen-category generalization in the industrial and clinical domains, and matches the strongest reference on macro-average root-cause accuracy. The approach applies where a deterministic step verifier exists, itself a real engineering investment, and broader claims await broader oracles; within that scope, an executable specification, not the model's own judgment, keeps evolution grounded.

\bibliography{references}

@article{singhal2023large,
  author  = {Singhal, Karan and Azizi, Shekoofeh and Tu, Tao and Mahdavi, S. Sara and Wei, Jason and Chung, Hyung Won and Scales, Nathan and Tanwani, Ajay and Cole-Lewis, Heather and Pfohl, Stephen and others},
  title   = {Large Language Models Encode Clinical Knowledge},
  journal = {Nature},
  volume  = {620},
  number  = {7972},
  pages   = {172--180},
  year    = {2023}
}

@inproceedings{guha2023legalbench,
  author    = {Guha, Neel and Nyarko, Julian and Ho, Daniel E. and R{\'e}, Christopher and Chilton, Adam and Narayana, Aditya and Chohlas-Wood, Alex and Peters, Austin and Waldon, Brandon and Rockmore, Daniel N. and others},
  title     = {LegalBench: A Collaboratively Built Benchmark for Measuring Legal Reasoning in Large Language Models},
  booktitle = {Advances in Neural Information Processing Systems},
  volume    = {36},
  year      = {2023}
}

@article{appliedenergy2024fault,
  author  = {Zhang, Jian and Zhang, Chaobo and Lu, Jie and Zhao, Yang},
  title   = {Domain-Specific Large Language Models for Fault Diagnosis of Heating, Ventilation, and Air Conditioning Systems by Labeled-Data-Supervised Fine-Tuning},
  journal = {Applied Energy},
  volume  = {377},
  pages   = {124378},
  year    = {2025}
}

@article{hinton2015distilling,
  author  = {Hinton, Geoffrey and Vinyals, Oriol and Dean, Jeff},
  title   = {Distilling the Knowledge in a Neural Network},
  journal = {arXiv preprint arXiv:1503.02531},
  year    = {2015}
}

@inproceedings{gu2024minillm,
  author    = {Gu, Yuxian and Dong, Li and Wei, Furu and Huang, Minlie},
  title     = {MiniLLM: Knowledge Distillation of Large Language Models},
  booktitle = {The Twelfth International Conference on Learning Representations},
  year      = {2024}
}

@inproceedings{agarwal2024onpolicy,
  author    = {Agarwal, Rishabh and Vieillard, Nino and Zhou, Yongchao and Stanczyk, Piotr and Ramos, Sabela and Geist, Matthieu and Bachem, Olivier},
  title     = {On-Policy Distillation of Language Models: Learning from Self-Generated Mistakes},
  booktitle = {The Twelfth International Conference on Learning Representations},
  year      = {2024}
}

@article{shao2024deepseekmath,
  author  = {Shao, Zhihong and Wang, Peiyi and Zhu, Qihao and Xu, Runxin and Song, Junxiao and Bi, Xiao and Zhang, Haowei and Zhang, Mingchuan and Li, Y. K. and Wu, Y. and Guo, Daya},
  title   = {DeepSeekMath: Pushing the Limits of Mathematical Reasoning in Open Language Models},
  journal = {arXiv preprint arXiv:2402.03300},
  year    = {2024}
}

@inproceedings{zelikman2022star,
  author    = {Zelikman, Eric and Wu, Yuhuai and Mu, Jesse and Goodman, Noah D.},
  title     = {{STaR}: Bootstrapping Reasoning with Reasoning},
  booktitle = {Advances in Neural Information Processing Systems},
  volume    = {35},
  pages     = {15476--15488},
  year      = {2022}
}

@article{huang2025rzero,
  author  = {Huang, Chengsong and Yu, Wenhao and Wang, Xiaoyang and Zhang, Hongming and Li, Zongxia and Li, Ruosen and Huang, Jiaxin and Mi, Haitao and Yu, Dong},
  title   = {{R-Zero}: Self-Evolving Reasoning {LLM} from Zero Data},
  journal = {arXiv preprint arXiv:2508.05004},
  year    = {2025}
}

@inproceedings{zhao2025absolute,
  author    = {Zhao, Andrew and Wu, Yiran and Yue, Yang and Wu, Tong and Xu, Quentin and Yue, Yang and Lin, Matthieu and Wang, Shenzhi and Wu, Qingyun and Zheng, Zilong and Huang, Gao},
  title     = {Absolute Zero: Reinforced Self-play Reasoning with Zero Data},
  booktitle = {Advances in Neural Information Processing Systems},
  year      = {2025}
}

@inproceedings{yuan2024selfrewarding,
  author    = {Yuan, Weizhe and Pang, Richard Yuanzhe and Cho, Kyunghyun and Li, Xian and Sukhbaatar, Sainbayar and Xu, Jing and Weston, Jason},
  title     = {Self-Rewarding Language Models},
  booktitle = {Proceedings of the 41st International Conference on Machine Learning},
  year      = {2024}
}

@article{guo2025deepseekr1,
  author  = {Guo, Daya and Yang, Dejian and Zhang, Haowei and Song, Junxiao and Wang, Peiyi and Zhu, Qihao and Xu, Runxin and Zhang, Ruoyu and Ma, Shirong and Bi, Xiao and others},
  title   = {DeepSeek-R1 Incentivizes Reasoning in {LLMs} through Reinforcement Learning},
  journal = {Nature},
  volume  = {645},
  number  = {8081},
  pages   = {633--638},
  year    = {2025}
}

@article{chen2025mae,
  author  = {Chen, Yixing and Wang, Yiding and Zhu, Siqi and Yu, Haofei and Feng, Tao and Zhang, Muhan and Patwary, Mostofa and You, Jiaxuan},
  title   = {Multi-Agent Evolve: {LLM} Self-Improve through Co-evolution},
  journal = {arXiv preprint arXiv:2510.23595},
  year    = {2025}
}

@article{arnesen2024debate,
  author  = {Arnesen, Samuel and Rein, David and Michael, Julian},
  title   = {Training Language Models to Win Debates with Self-Play Improves Judge Accuracy},
  journal = {arXiv preprint arXiv:2409.16636},
  year    = {2024}
}

@article{chen2025sec,
  author  = {Chen, Xiaoyin and Lu, Jiarui and Kim, Minsu and Zhang, Dinghuai and Tang, Jian and Pich{\'e}, Alexandre and Gontier, Nicolas and Bengio, Yoshua and Kamalloo, Ehsan},
  title   = {Self-Evolving Curriculum for {LLM} Reasoning},
  journal = {arXiv preprint arXiv:2505.14970},
  year    = {2025}
}

@article{irving2018debate,
  author  = {Irving, Geoffrey and Christiano, Paul and Amodei, Dario},
  title   = {{AI} Safety via Debate},
  journal = {arXiv preprint arXiv:1805.00899},
  year    = {2018}
}

@inproceedings{du2024improving,
  author    = {Du, Yilun and Li, Shuang and Torralba, Antonio and Tenenbaum, Joshua B. and Mordatch, Igor},
  title     = {Improving Factuality and Reasoning in Language Models through Multiagent Debate},
  booktitle = {Proceedings of the 41st International Conference on Machine Learning},
  year      = {2024}
}

@inproceedings{khan2024debating,
  author    = {Khan, Akbir and Hughes, John and Valentine, Dan and Ruis, Laura and Sachan, Kshitij and Radhakrishnan, Ansh and Grefenstette, Edward and Bowman, Samuel R. and others},
  title     = {Debating with More Persuasive {LLMs} Leads to More Truthful Answers},
  booktitle = {Proceedings of the 41st International Conference on Machine Learning},
  year      = {2024}
}

@article{geirhos2020shortcut,
  author  = {Geirhos, Robert and Jacobsen, J{\"o}rn-Henrik and Michaelis, Claudio and Zemel, Richard and Brendel, Wieland and Bethge, Matthias and Wichmann, Felix A.},
  title   = {Shortcut Learning in Deep Neural Networks},
  journal = {Nature Machine Intelligence},
  volume  = {2},
  number  = {11},
  pages   = {665--673},
  year    = {2020}
}

@inproceedings{huang2024large,
  author    = {Huang, Jie and Chen, Xinyun and Mishra, Swaroop and Zheng, Huaixiu Steven and Yu, Adams Wei and Song, Xinying and Zhou, Denny},
  title     = {Large Language Models Cannot Self-Correct Reasoning Yet},
  booktitle = {Proceedings of the 12th International Conference on Learning Representations (ICLR)},
  year      = {2024}
}

@inproceedings{wang2024mathshepherd,
  author    = {Wang, Peiyi and Li, Lei and Shao, Zhihong and Xu, Runxin and Dai, Damai and Li, Yifei and Chen, Deli and Wu, Yu and Sui, Zhifang},
  title     = {Math-Shepherd: Verify and Reinforce {LLMs} Step-by-Step without Human Annotations},
  booktitle = {Proceedings of the 62nd Annual Meeting of the Association for Computational Linguistics},
  pages     = {9426--9439},
  year      = {2024}
}

@inproceedings{lightman2024lets,
  author    = {Lightman, Hunter and Kosaraju, Vineet and Burda, Yura and Edwards, Harri and Baker, Bowen and Lee, Teddy and Leike, Jan and Schulman, John and Sutskever, Ilya and Cobbe, Karl},
  title     = {Let's Verify Step by Step},
  booktitle = {The Twelfth International Conference on Learning Representations},
  year      = {2024}
}

@article{uesato2022solving,
  author  = {Uesato, Jonathan and Kushman, Nate and Kumar, Ramana and Song, Francis and Siegel, Noah and Wang, Lisa and Creswell, Antonia and Irving, Geoffrey and Higgins, Irina},
  title   = {Solving Math Word Problems with Process- and Outcome-Based Feedback},
  journal = {arXiv preprint arXiv:2211.14275},
  year    = {2022}
}

@inproceedings{zhang2025lessons,
  author    = {Zhang, Zhenru and Zheng, Chujie and Wu, Yangzhen and Zhang, Beichen and Lin, Runji and Yu, Bowen and Liu, Dayiheng and Zhou, Jingren and Lin, Junyang},
  title     = {The Lessons of Developing Process Reward Models in Mathematical Reasoning},
  booktitle = {Findings of the Association for Computational Linguistics: ACL 2025},
  year      = {2025}
}

@inproceedings{wei2022chain,
  author    = {Wei, Jason and Wang, Xuezhi and Schuurmans, Dale and Bosma, Maarten and Ichter, Brian and Xia, Fei and Chi, Ed and Le, Quoc V. and Zhou, Denny},
  title     = {Chain-of-Thought Prompting Elicits Reasoning in Large Language Models},
  booktitle = {Advances in Neural Information Processing Systems},
  volume    = {35},
  pages     = {24824--24837},
  year      = {2022}
}

@inproceedings{turpin2023language,
  author    = {Turpin, Miles and Michael, Julian and Perez, Ethan and Bowman, Samuel R.},
  title     = {Language Models Don't Always Say What They Think: Unfaithful Explanations in Chain-of-Thought Prompting},
  booktitle = {Advances in Neural Information Processing Systems},
  volume    = {36},
  year      = {2023}
}

@inproceedings{gudibande2024imitation,
  author    = {Gudibande, Arnav and Wallace, Eric and Snell, Charlie and Geng, Xinyang and Liu, Hao and Abbeel, Pieter and Levine, Sergey and Song, Dawn},
  title     = {The False Promise of Imitating Proprietary {LLMs}},
  booktitle = {The Twelfth International Conference on Learning Representations},
  year      = {2024}
}

@inproceedings{wang2023selfinstruct,
  author    = {Wang, Yizhong and Kordi, Yeganeh and Mishra, Swaroop and Liu, Alisa and Smith, Noah A. and Khashabi, Daniel and Hajishirzi, Hannaneh},
  title     = {Self-Instruct: Aligning Language Models with Self-Generated Instructions},
  booktitle = {Proceedings of the 61st Annual Meeting of the Association for Computational Linguistics},
  pages     = {13484--13508},
  year      = {2023}
}

@inproceedings{sukhbaatar2018intrinsic,
  author    = {Sukhbaatar, Sainbayar and Lin, Zeming and Kostrikov, Ilya and Synnaeve, Gabriel and Szlam, Arthur and Fergus, Rob},
  title     = {Intrinsic Motivation and Automatic Curricula via Asymmetric Self-Play},
  booktitle = {Proceedings of the 6th International Conference on Learning Representations (ICLR)},
  year      = {2018}
}

@inproceedings{bengio2009curriculum,
  author    = {Bengio, Yoshua and Louradour, Jerome and Collobert, Ronan and Weston, Jason},
  title     = {Curriculum Learning},
  booktitle = {Proceedings of the 26th International Conference on Machine Learning (ICML)},
  year      = {2009}
}

@techreport{wen2011rp1312,
  author      = {Wen, Jin and Li, Shun},
  title       = {{RP-1312} -- Tools for Evaluating Fault Detection and Diagnostic Methods for Air-Handling Units},
  institution = {ASHRAE},
  year        = {2011}
}

@article{yu2025mslr,
  author  = {Yu, Wenhan and Lin, Xinbo and Ni, Lanxin and Cheng, Jinhua and Sha, Lei},
  title   = {Benchmarking Multi-Step Legal Reasoning and Analyzing Chain-of-Thought Effects in Large Language Models},
  journal = {arXiv preprint arXiv:2511.07979},
  year    = {2025}
}

@misc{granderson2022lbnl,
  author       = {Granderson, Jessica and Lin, Guanjing and Chen, Yimin and Casillas, Armando and Im, Piljae and Jung, Sungkyun and Benne, Kyle and Ling, Jiazhen and Gorthala, Ravi and Wen, Jin and Chen, Zhelun and Huang, Sen and Vrabie, Draguna},
  title        = {{LBNL} Fault Detection and Diagnostics Datasets},
  year         = {2022},
  howpublished = {Open Energy Data Initiative (OEDI), Lawrence Berkeley National Laboratory},
  doi          = {10.25984/1881324}
}

@inproceedings{tchango2022ddxplus,
  author    = {Fansi Tchango, Ars{\`e}ne and Goel, Rishab and Wen, Zhi and Martel, Julien and Ghosn, Joumana},
  title     = {{DDXPlus}: A New Dataset for Automatic Medical Diagnosis},
  booktitle = {Advances in Neural Information Processing Systems 35, Datasets and Benchmarks Track},
  year      = {2022}
}

@inproceedings{pan2023logiclm,
  author    = {Pan, Liangming and Albalak, Alon and Wang, Xinyi and Wang, William},
  title     = {{Logic-LM}: Empowering Large Language Models with Symbolic Solvers for Faithful Logical Reasoning},
  booktitle = {Findings of the Association for Computational Linguistics: EMNLP 2023},
  pages     = {3806--3824},
  year      = {2023}
}

@article{kamoi2024when,
  author    = {Kamoi, Ryo and Zhang, Yusen and Zhang, Nan and Han, Jiawei and Zhang, Rui},
  title     = {When Can {LLMs} Actually Correct Their Own Mistakes? {A} Critical Survey of Self-Correction of {LLMs}},
  journal   = {Transactions of the Association for Computational Linguistics},
  volume    = {12},
  pages     = {1417--1440},
  year      = {2024}
}

@inproceedings{jin2023cladder,
  author    = {Jin, Zhijing and Chen, Yuen and Leeb, Felix and Gresele, Luigi and Kamal, Ojasv and Lyu, Zhiheng and Blin, Kevin and Gonzalez Adauto, Fernando and Kleiman-Weiner, Max and Sachan, Mrinmaya and Sch{\"o}lkopf, Bernhard},
  title     = {{CLadder}: Assessing Causal Reasoning in Language Models},
  booktitle = {Advances in Neural Information Processing Systems},
  volume    = {36},
  year      = {2023}
}

@article{bodenreider2004umls,
  author  = {Bodenreider, Olivier},
  title   = {The Unified Medical Language System ({UMLS}): Integrating Biomedical Terminology},
  journal = {Nucleic Acids Research},
  volume  = {32},
  number  = {suppl\_1},
  pages   = {D267--D270},
  year    = {2004}
}

\end{document}